\documentclass[conference]{IEEEtran}
\IEEEoverridecommandlockouts
\usepackage{balance}

\usepackage{amsmath,amssymb,amsfonts}
\usepackage{graphicx}
\usepackage{textcomp}
\usepackage{xcolor}
\usepackage{hyperref}

\usepackage{float} 
\def\BibTeX{{\rm B\kern-.05em{\sc i\kern-.025em b}\kern-.08em
    T\kern-.1667em\lower.7ex\hbox{E}\kern-.125emX}}
\usepackage[ruled, linesnumbered, commentsnumbered, longend]{algorithm2e}
\usepackage{array} 

\usepackage{amsmath}
\usepackage{amsfonts}
\usepackage{mathtools}
\usepackage{bm}

\newcommand{\HS}{\mathcal{H}} 

\def\vh{{\bm{h}}}

\begin{document}

\title{Enhanced Detection of Transdermal Alcohol Levels Using Hyperdimensional Computing on Embedded Devices}

\newcommand{\linebreakand}{%
  \end{@IEEEauthorhalign}
  \hfill\mbox{}\par
  \mbox{}\hfill\begin{@IEEEauthorhalign}
}





\author{\IEEEauthorblockN{Manuel E. Segura\textsuperscript{\textsection}}
\IEEEauthorblockA{
\textit{University of California, Irvine}\\
Irvine, USA \\
mesegur1@uci.edu}
\and
\IEEEauthorblockN{Pere Vergés\textsuperscript{\textsection}}
\IEEEauthorblockA{
\textit{University of California, Irvine}\\
Irvine, USA \\
pvergesb@uci.edu}
\and
\IEEEauthorblockN{Justin Tian Jin Chen}
\IEEEauthorblockA{
\textit{University of California, Irvine}\\
Irvine, USA \\
chenjt3@uci.edu}
\and
\IEEEauthorblockN{Ramesh Arangott}
\IEEEauthorblockA{
\textit{Asahi Group Holdings Ltd.}\\
Tokyo, Japan \\
ramesh.arangott@asahigroup-holdings.com}
\and
\IEEEauthorblockN{Angela Kristine Garcia}
\IEEEauthorblockA{
\textit{Asahi Group Holdings Ltd.}\\
Tokyo, Japan \\
angela.kristinegarcia@asahigroup-holdings.com}
\and
\IEEEauthorblockN{Laura Garcia Reynoso}
\IEEEauthorblockA{
\textit{Asahi Group Holdings Ltd.}\\
Tokyo, Japan \\
laura.garcia@asahigroup-holdings.com}
\and
\IEEEauthorblockN{Alex Nicolau}
\IEEEauthorblockA{
\textit{University of California, Irvine}\\
Irvine, USA \\
nicolau@ics.uci.edu}
\and
\IEEEauthorblockN{Tony Givargis}
\IEEEauthorblockA{
\textit{University of California, Irvine}\\
Irvine, USA \\
givargis@uci.edu}
\and
\IEEEauthorblockN{Sergio Gago-Masague}
\IEEEauthorblockA{
\textit{University of California, Irvine}\\
Irvine, USA \\
sgagomas@uci.edu}
}

\author{\IEEEauthorblockN{Manuel E. Segura\textsuperscript{\textsection}\textsuperscript{*},
Pere Vergés\textsuperscript{\textsection}\textsuperscript{*}, Justin Tian Jin Chen\textsuperscript{*},
Ramesh Arangott\textsuperscript{+}, Angela Kristine Garcia\textsuperscript{+}, \\ 
Laura Garcia Reynoso\textsuperscript{+}, Alexandru Nicolau\textsuperscript{*}, Tony Givargis\textsuperscript{*} and Sergio Gago-Masague\textsuperscript{*}
}
\IEEEauthorblockA{University of California Irvine\textsuperscript{*}, Asahi Group Holdings Ltd.\textsuperscript{+}\\
mesegur1@uci.edu, pvergesb@uci.edu, chenjt3@uci.edu, ramesh.arangott@asahigroup-holdings.com, \\
angela.kristinegarcia@asahigroup-holdings.com, laura.garcia@asahigroup-holdings.com, \\
nicolau@ics.uci.edu, givargis@uci.edu, sgagomas@uci.edu}}

\maketitle
\begingroup\renewcommand\thefootnote{\textsection}
\footnotetext{Both authors contributed equally to this research.}

\begin{abstract}
Alcohol consumption has a significant impact on individuals' health, with even more pronounced consequences when consumption becomes excessive. One approach to promoting healthier drinking habits is implementing just-in-time interventions, where timely notifications indicating intoxication are sent during heavy drinking episodes. However, the complexity or invasiveness of an intervention mechanism may deter an individual from using them in practice. Previous research tackled this challenge using collected motion data and conventional Machine Learning (ML) algorithms to classify heavy drinking episodes, but with impractical accuracy and computational efficiency for mobile devices. Consequently, we have elected to use Hyperdimensional Computing (HDC) to design a just-in-time intervention approach that is practical for smartphones, smart wearables, and IoT deployment. HDC is a framework that has proven results in processing real-time sensor data efficiently. This approach offers several advantages, including low latency, minimal power consumption, and high parallelism. We explore various HDC encoding designs and combine them with various HDC learning models to create an optimal and feasible approach for mobile devices. Our findings indicate an accuracy rate of 89\%, which represents a substantial 12\% improvement over the current state-of-the-art.
\end{abstract}



\maketitle


\section{Introduction}
Alcohol is a harmful substance that contributes to more than 200 diseases, injuries, and health conditions~\cite{alcohol}. According to the World Health Organization, in 2022, 5.3\% of global deaths were attributed to alcohol use, accounting for three million deaths annually. Furthermore, alcohol consumption is known to lead to various mental and behavioral disorders and injuries. The risks associated with alcohol significantly escalate when excessive consumption occurs within short time intervals~\cite{alcoholconsumption}.

To address this concern, a pivotal effort is aimed at minimizing associated risks. Various strategies can be deployed, such as educating the general population and implementing screenings and interventions. In our study, we aim to alert individuals when they approach the legal blood alcohol limit, typically set at 0.08\%\cite{BAClimit} for those aged 21 and above—this benchmark guides our investigation. However, studies suggest that lowering the blood alcohol content (BAC) level to 0.05\%\cite{BAClimit2} could reduce the occurrence of fatal crashes involving impaired driving. By adopting this lower threshold, individuals can make informed choices, diminishing the likelihood of excessive alcohol consumption and the risk of driving under the influence. Providing users with a tool to assess their alcohol levels can heighten awareness and foster responsible drinking habits.

Various methods exist for measuring BAC, such as breathalyzers, blood, urine, and saliva tests. However, these approaches can be intrusive and require user cooperation. Another option, transdermal alcohol concentration (TAC), involves using specific wrist or ankle strap devices for measurements. This study aims to establish a strong correlation between TAC measurements and accelerometer data from users' smartphones, achieving accurate results without intruding on the user experience. This is particularly significant given the widespread use of smartphones and the inherent presence of accelerometers in these devices.

While prior investigations have delved into this subject~\cite{Killian2019LearningTD,bac2}, our objective is to enhance both accuracy and efficiency. Previous studies predominantly leaned on conventional machine learning algorithms, including Support Vector Machines (SVM), Convolutional Neural Networks (CNN), or Random Forest~\cite{Killian2019LearningTD,10.1145/3234150}, for classifying sobriety. Yet, the intricacy of this challenge lies in accurately determining sobriety across diverse scenarios, such as when the smartphone is in a pocket, on a table, or actively in use (e.g., gaming, calling, texting, or using various applications). Moreover, the data may be prone to noise, and signals may intermittently be lost, presenting hurdles in solving this issue. Our aspiration is to develop a less resource-intensive solution that can operate on or with smartwearables such as smartwatches or smart bracelets, enabling the capture of information in a more seamless manner.

Hence, our approach involves leveraging Hyperdimensional Computing (HDC) for classification. HDC has demonstrated impressive results in handling noisy and real-time data across different applications, including Electromyography (EMG)~\cite{8919214}, Electroencephalography (EEG), and Electrocorticography (ECoG). This technology provides several advantages, including low latency, low power consumption, and high parallelism, making it an excellent framework for addressing challenges within the Internet of Things (IoT) domain.

\section{Related work}
Numerous prior studies have examined the area of predicting blood alcohol content in the human body. One notable approach stemmed from a mathematical model introduced by E.M.P. Widmark in 1932 (further improvements of the equation have been proposed~\cite{bloodAlcoholWidmark}), known as the Widmark equation. 
Subsequent approaches in this domain have also emerged, including transdermal ethanol sensors for TAC detection. These sensors are designed to gauge alcohol levels by detecting the gaseous phase of alcohol in the air surrounding the skin. These studies~\cite{tac1,tac2} have revealed that TAC can accurately identify moderate alcohol consumption, reflected when TAC levels range between 0 and 0.02 g/dl. Nevertheless, they are not reliable indicators of lower drinking levels (i.e. whether an individual has ceased drinking), unless the number of drinks consumed is known~\cite{ROACHE2019101}.

The proliferation of smartphones and other mobile technologies in the last 20 years have enabled research in the field of just-in-time adaptive interventions (JITAI). The purpose of JITAIs are to provide personalized real-time interventions for users, typically with a mobile device, using data collected from the mobile device~\cite{Reichert2020-eq}. These have proven highly effective at helping with issues such as weight management~\cite{Patrick2009-ca}, managing stress~\cite{stress}, exercise~\cite{exercise}, nicotine addiction~\cite{Riley2008-hj}, and alcohol consumption~\cite{Gustafson2014-lf, Nahum-Shani2018-ln}. In particular, the study performed in \cite{Gustafson2014-lf} demonstrated that personalized JITAIs deterred hazardous behavior during drinking episodes.

The application of mathematical models has led to less intrusive methods for predicting BAC or TAC, thereby opening the door for predictive algorithms to be used JITAIs. Another study~\cite{tac3}, employed a mathematical model that incorporates TAC data and the specific types of beverages an individual intends to consume to predict BAC. Furthermore, novel techniques have emerged that utilize machine learning and mobile data, including accelerometer data, to provide a cost-effective and practical way of predicting BAC or TAC. These endeavors deploy classification models, such as Random Forest~\cite{bac2,Killian2019LearningTD}, CNNs~\cite{bac3,Killian2019LearningTD}, or SVMs~\cite{bac1,Killian2019LearningTD}, to harness sensor data for accurate predictions. Similarly, regression models have also been explored. Bayesian regression neural networks (BRNN) have been used to perform regression analysis from accelerometer data~\cite{bac3}, and furthermore, an ensemble of CNNs and bi-directional Long Short Term Memory (bi-LSTM) were also used for regression analysis that yielded promising results~\cite{bac4}. However, these neural networks are often computationally expensive to train and perform inference with, so they were not in the scope of our work for exploring practical methods for real-time IoT applications for JITAIs. Moreover, there was no code publicly available to reproduce the claimed results.

Remarkably, HDC, which has demonstrated notable success in embedded systems and time-series data, has not been previously employed in classification within this domain. Its effectiveness is evident through outstanding outcomes characterized by a combination of low energy consumption and high accuracy across a diverse array of applications. These applications span from emotion recognition~\cite{8771622} and hand gesture recognition~\cite{Moin2020AWB,8919214,7738683} to language recognition~\cite{10.1145/2934583.2934624} and multiclass classification of electromyography (EMG), electroencephalography (EEG), and electrocorticography (ECoG)~\cite{8490896} biosignals. As a result, our HDC solution is strategically positioned to deliver JITAIs with effective performance on smartphones and smart wearables.
\section{Hyperdimensional Computing}
HDC, a brain-inspired framework, relies on high-dimensional vectors to represent information. These hypervectors have a dimensionality, ($d$) on the order of thousands which are randomly sampled from hyperspace. Depending on the application's requirements, a range of hyperspaces can be selected~\cite{plate2003holographic,gayler1998multiplicative,plate1995holographic,kanerva1997fully}. These spaces encompass a spectrum from binary to complex value representations. Among them, the Multiply Add Permute (MAP)~\cite{gayler1998multiplicative} hyperspace is one of the most popular thanks to its efficiency and information representation capabilities, for that reason we have selected for our application.

\subsection{Distance Metric}
Measuring the proximity of two hypervectors is crucial due to their high dimensionality and the importance of orthogonality. This metric determines whether two hypervectors are identical, similar, or vastly dissimilar. In our context, this metric aids in classifying which class a particular hypervector belongs to, facilitating the accurate classification of data points. Typically, the distance metric is computed using the dot product, although, depending on data characteristics and normalization, cosine similarity or hamming distance may also be applied.

\subsection{Operations}
HDC relies on three fundamental operations:

\paragraph{\textbf{Superposition}} This operation, also known as bundling, aggregates information from various hypervectors into a single hypervector. It can be represented as $\oplus: \HS\times\HS\to\HS$. Notably, the resulting hypervector bears a high resemblance to each hypervector contributing to its formation. This operation is typically implemented as element-wise addition, rendering it high efficiency and parallelizability~\cite{plate2003holographic,kanerva1988sparse}.

\paragraph{\textbf{Binding}}
This operation binds two variables together, similar to traditional computing's assignment of a value to a variable. It is represented as $\otimes: \HS \times \HS \to \HS$. Unlike superposition, the resulting hypervector is entirely distinct from its components. In vector symbolic architectures, the inverse operation of binding can be performed by binding the variable again with the same hypervector, effectively restoring it to its original state, therefore making the binding operation its self-inverse.

\paragraph{\textbf{Permute}} 
This operation establishes order among hypervectors, this is vital for creating data structures and capturing data patterns in time series applications~\cite{8919214}. It can be represented as $p^{i}(\vh)$ where $i$ denotes the number of times the operation is applied. Similar to binding, the resulting hypervector after the operation is dissimilar. The inverse permute operation, $p^{-i}(\vh),$ reverts the hypervector to its exact original state~\cite{cohen-widdows-2018-bringing, EasyChair:11190}.

\subsection{Hyperdimensional Classification}
The HDC framework has incited interest in solving various machine learning problems, spanning from regression~\cite{9586284}, reinforcement learning~\cite{10.1145/3508352.3549437}, and classification tasks~\cite{vergés2023hdcc}~\cite{HDC_classification}. The fundamental workflow for hyperdimensional learning includes encoding training samples into the hyperspace, adding each sample to its respective class, and storing this information in associative memory, as illustrated in Figure~\ref{fig:workflow}.

\begin{figure}[htbp!]
\centering
\includegraphics[width=6cm]{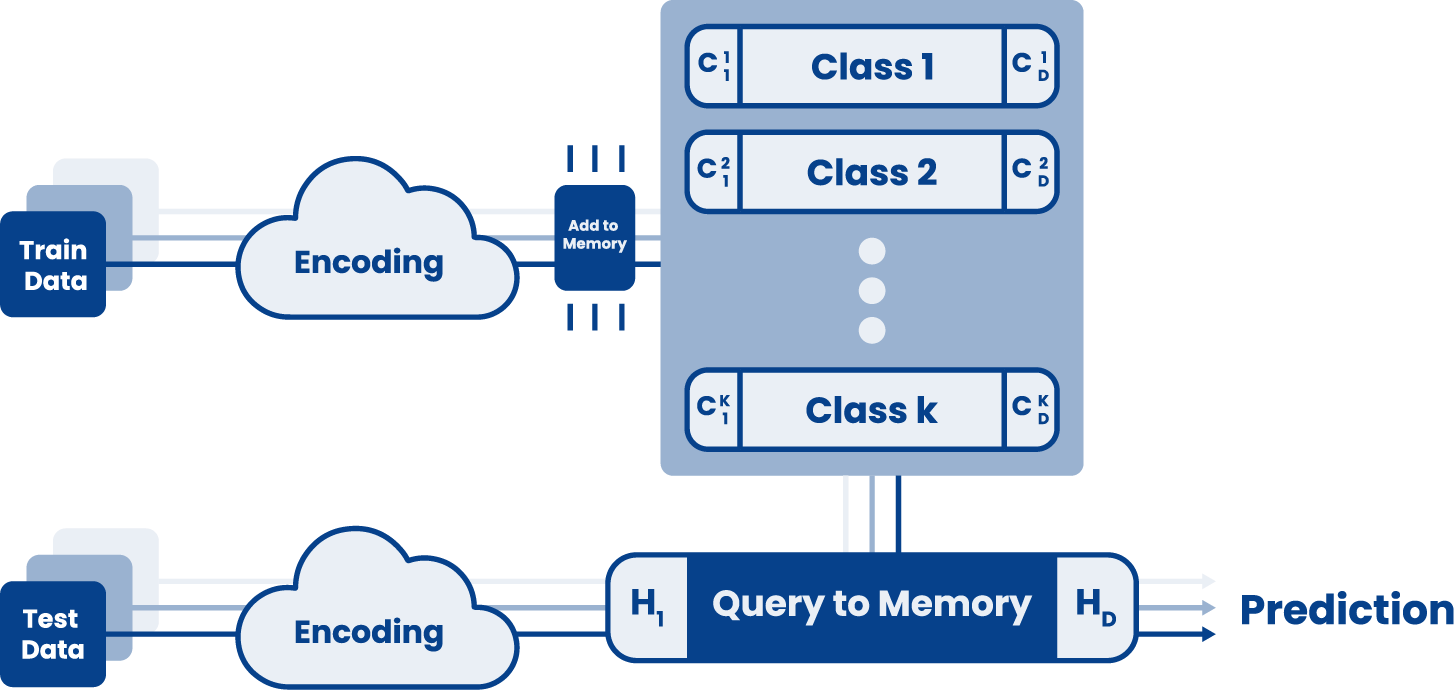}
 \caption{\label{fig:workflow} Hyperdimensional classification workflow.}
\end{figure}

Once the training phase concludes, inference begins, encoding test samples to the hyperspace, where it's important to note that the encoding process remains consistent for both training and inference. For each encoded sample, a similarity metric is employed to determine the class most similar to the given instance, ultimately facilitating classification. While this serves as the fundamental workflow, several enhancements to this approach have been explored. These enhancements range from improved single-pass learning using adaptive learning~\cite{8918974,9474107}, to iterative training~\cite{9474107}, and the use of dimension regeneration~\cite{zou2021scalable,wang2023disthd}, for dimensions carrying minimal to no information.

\subsection{Encodings}\label{Encodings}
The encoding phase within the hyperdimensional learning workflow is computationally intensive and significantly impacts classification outcomes. Encodings can be either unstructured or structured. Unstructured encodings are typically created through vector projections, including random projection~\cite{Thomas2021}, sinusoid projection~\cite{zou2021scalable}, or fractional power encoding~\cite{frady2021computing,article}. Structured encodings employ different types of basis hypervectors, such as random hypervectors, level hypervectors with linear correlations, and circular hypervectors suitable for circular data representation, such as seasonal variations. These basis hypervectors are utilized to create various structures, such as the key-value encoding~\cite{Ge2020,9388914,Thomas2021}, n-gram encoding~\cite{6868349} and the generic encoding~\cite{10.1145/3489517.3530669}, which combines elements of both n-gram and key-value encoding. Moreover one can create more specific encodings that better tailor its application by combining the aforementioned encodings using hyperdimensional operations.

\section{Methods}
The heavy drinking classification problem has been traditionally addressed through classical machine learning (ML) techniques in prior studies. However, our approach diverged as we opted for HDC, leveraging its exceptional performance in both embedded systems and classification tasks~\cite{8919214}. Our objective was to utilize the high parallelism, low latency, and low power consumption of HDC to develop a solution that runs efficiently on IoT devices like smart wearables and smartphones, minimizing resource contention. The initial stage of our method focused on data treatment, which is a critical step for this application. This was particularly significant because the problem is meant to be addressed in real-time. It was also essential to have a streamlined and efficient method for processing input data, along with the removal of noise. Providing accurate parameters to the HDC classification model is extremely important for obtaining high accuracy results.

\subsection{Data Selection and Cleaning}
We utilized the "Bar Crawl: Detecting Heavy Drinking" dataset from a previous study~\cite{Killian2019LearningTD, misc_bar_crawl:_detecting_heavy_drinking_515}. This dataset, gathered from 11 iOS and 2 Android smartphones and TAC measurements from a SCRAM ankle bracelet, consists of a real-time series with a substantial 14 million accelerometer readings and 715 TAC (Transdermal Alcohol Concentration) readings. The usable data for the analysis consists of the reading of 13 participants~\cite{Killian2019LearningTD}. The raw accelerometer data was collected at a sampling rate of 40Hz~\cite{Killian2019LearningTD}. The TAC data, obtained at 30-minute intervals using SCRAM ankle bracelets, underwent two processing steps: (1) a zero-phase low-pass filter to reduce noise without shifting phase, and (2) a backward shift of 45 minutes to align labels more closely with the true intoxication of the participant, considering the 45-minute duration for alcohol to exit through the skin~\cite{Killian2019LearningTD}. Subsequently, the raw accelerometer data was windowed into 10-second overlapping windows (400 samples per window), with TAC data interpolated for those 10-second intervals~\cite{Killian2019LearningTD}.

Minimizing the amount of data and treatment done to the data is fundamental to reducing energy and execution time in a real-time application, where fast and efficient responses are necessary to be able to execute the application in a real scenario. We considered the same engineered feature types as used in a previous study~\cite{Killian2019LearningTD}, which resulted in a total of 600 features. Similar to the previous study, we used Autogluon~\cite{erickson2020autogluontabular} to identify the top 20\% (120) of features with the highest importance scores, which measure how much each feature influenced the resulting prediction. We then used these 120 features in our approach. However, upon further examination, only 18 individual features had importance higher than 0.0001, making clear that most of the 600 features almost did not contribute to the right prediction. The feature types we considered are listed in Table~\ref{tab:feat_selection}. The list of types corroborated the importance of Mel Frequency Cepstral Coefficients (MFCC) for this problem from the previous study~\cite{Killian2019LearningTD}.

\begin{table}[!ht]
\centering
\footnotesize
\caption{Categories of most relevant features considered per window. \label{tab:feat_selection}}
\begin{tabular}{|l|m{5.5cm}|}
    \hline
    \textbf{Feature Type} & \textbf{Description}  \\ \hline
    Mean & Average of each x, y, z signal  \\ \hline
    Standard Deviation  & Standard deviation of each x, y, z signal \\ \hline
    Median              & Median of each x, y, z signal \\ \hline
    Spectral Entropy    & Entropy in frequency and time domain (2 total metrics) for each axis\\ \hline
    Spectral Centroid   & Weighted mean of frequencies for each axis \\ \hline
    Spectral Spread     & Measure of variance about the centroid for each axis\\ \hline
    RMS                 & Root-mean-square of accelerations for each axis \\ \hline
    MFCC Covariance     & MFCC covariance entries for each of 6 axis combinations  \\ \hline
\end{tabular}
\end{table}

\subsection{HDC Approach}\label{HDC Approach}

With an HDC approach, we first determined how data will be encoded. In the literature~\cite{HDC_classification} we observed that there is a certain set of general encodings that have proved to work well in similar applications. The four encodings mentioned in Section~\ref{Encodings} that we used for raw accelerometer time-series data were the following:
\textit{Key-Value (KV)~\cite{Ge2020,9388914,Thomas2021}}, \textit{N-gram (N)~\cite{6868349}}, \textit{Generic~\cite{10.1145/3489517.3530669}}, \textit{Sinusoid projection (S. Proj)~\cite{zou2021scalable}}.

We used the \textit{Density} encoding~\cite{9174774} for the engineered feature data. This encoding is an alternate implementation of level hypervectors to encode information. Our final ensemble encoding method was a combination of the Density encoding and one of the four mentioned previously. 

In addition to developing an encoding method, we needed to determine which learning method model would work best for our classification problem. Several works have proposed different classification learning models. These models differ in accuracy and training time. However, inference time is very similar for all of them~\cite{HDC_classification}. We considered 6 different models:
\begin{itemize}
    \item \textit{VanillaHD (V)}, proposes an approach where all samples are added to the memory without modification. This is a single-pass algorithm.
    \item \textit{AdaptHD (A)~\cite{8918974}}, This model performs adaptive learning, this is done by adding a sample to the memory if it is misclassified. Can be single-pass or perform iterative learning.
    \item \textit{OnlineHD (O)~\cite{9474107}}, This model builds upon \textit{AdaptHD} by doing a weighted addition of the samples. It can be single-pass or perform iterative learning.
    \item \textit{RefineHD (R)~\cite{EasyChair:11191}}, this model improves upon \textit{OnlineHD} and \textit{AdaptHD} by adding samples that are correctly classified but differ largely from the class. Moreover, it proposes the use of the flocet encoding in the key-value encoding. It can be single-pass or perform iterative learning.
    \item \textit{NeuralHD (N)~\cite{zou2021scalable}}, this model uses dimensions regeneration iterative training, and adaptive learning to improve the model's learning.
    \item \textit{DistHD (D)~\cite{wang2023disthd}}, this model combines dimension regeneration, top two classifications, adaptive learning, and iterative training to achieve better accuracy
\end{itemize}
From these models, the first four can be performed using single-pass learning which is extremely useful for fast and low-resource training and inference. If the training can be offloaded and performed asynchronously, then performing iterative training can help achieve better results.

The HDC learning method models offer a key advantage by allowing the capture of temporal information while training the model. We leveraged this advantage by using the 400-sample raw time-series data windows in our HDC approach, combining it with our engineered features for classification. This is reflected in our HDC encoding selection, where we explored HDC encodings used to encode time-series data in order to develop our own unique encoder. For this study, we used both raw data and feature-engineered data. We used two approaches for this problem, one where we shuffled the windows of all 13 participant and one where we used the windows sorted in chronological order, this second approach was thought to work better thanks to the inherent time dependence that this problem carries. For example, in the HDC implementation of the time-series classification algorithm MiniROCKET named HDC-MiniROCKET, timestamps were able to be encoded in addition to the feature data provided by the convolutions, which allowed for significant accuracy improvements ~\cite{schlegel2022hdcminirocket}. Similarly, we aimed to consider longer patterns occurring during the bar crawl data collection period to enhance overall accuracy.

\subsection{Outline of Experiments}
The outline of our experiments~\footnote{Project link: \url{https://anonymous.4open.science/r/bar\_crawl\_data-B645}} was split into 3 phases. The implementation of each phase of the experiments was done using the Torchhd~\cite{JMLR:v24:23-0300} a python library. The first phase consisted of comparing combinations of the \textit{Density} encoding with the 4 other encodings used for the raw time-series windows.  We used \textit{RefineHD} as a default for testing in this phase. This provided us with an optimal encoding method for the second phase, learning method model selection. In the second phase we used our finalized ensemble encoding method with our 6 potential learning method models to determine the best encoding and learning method combination. Our third and final phase of experiments consisted of refining our final encoding and learning method combination with hyperparameter refinement.

For the execution of our experiments, we employed a Raspberry Pi 4 Model B with a Quad core Cortex-A72 (ARM v8) and 4GB of LPDDR4-3200 SDRAM memory~\cite{RaspberryPi}. We chose this platform as it's computational specifications are similar in specifications to a low-end smartphone.

We divided the data into two sets, with a 70/30 ratio. When the data was randomly shuffled, we divided it randomly. For chronologically ordered data, we reserved the last 30\%. 
\section{Results and Discussion}

Our results are split into the three major phases of our experiments: encoding selection, learning method model selection, and hyperparameter tuning. We also performed a baseline comparison with the results of the previous study, and explore execution times of the different encodings on the Raspberry Pi hardware.

\subsection{First Phase Results - Encoding Selection}
Our findings for the encoding selection, presented in Table~\ref{tab:encoding_selection_s} and Table~\ref{tab:encoding_selection_o}, shed light on the encoding selection process. For each encoding that is paired with the \textit{Density} encoding, we recorded the accuracy and F1-score and compare results. The F1-scores were expected lower with the chronologically ordered data, due to TAC levels being lower in the chronologically last 30\% of the data (mostly the negative "sober" label). The \textit{Generic} encoding and \textit{Density} encoding combination attains the one of the highest accuracy levels for both shuffled and ordered data. The breakdown of our final ensemble encoder is shown in Algorithm~\ref{alg:encoder}.


\begin{algorithm}
\caption{Ensemble HDC Encoder}\label{alg:encoder}
\SetKwInOut{KwIn}{Input}
\SetKwInOut{KwOut}{Output}
\SetKwFunction{GenericEncode}{GenericEncode}
\SetKwFunction{DensityEncode}{DensityEncode}

\KwIn{A window $W$, composed of raw accelerometer data $A$, and engineered features $F$.}
\KwOut{A hypervector.}
    \tcp{Apply Generic encoding to x, y, and z axis to encode.}
    $H_1 \leftarrow \GenericEncode(A.time, A.x, A.y, A.z)$

    \tcp{Apply Density encoding to 120 engineered features.}
    $H_2 \leftarrow \DensityEncode(F)$

    \tcp{Bind and bundle $H_1$ and $H_2$.}
    $H_f \leftarrow H_1 \oplus H_2 \oplus (H_1 \otimes H_2)$
    
    \KwRet{$H_f$}
\end{algorithm}

\begin{table}[!ht]
    \centering
    \caption{Performance evaluation of the HDC encodings, with randomly shuffled data.\label{tab:encoding_selection_s}}

    \begin{tabular}{|l|l|l|l|l|}
    \hline
        \textbf{Metric} & \textbf{KV (RL)} & \textbf{S. Proj} & \textbf{KV (SN)} & \textbf{Generic} \\ \hline
        Accuracy  & 79.069 & 78.78 &  \textbf{82.857} & 82.41  \\ \hline
        F1-Score & 0.571 & 0.572 & 0.67 &\textbf{0.692}  \\ \hline
    \end{tabular}
\end{table}

\begin{table}[!ht]
    \centering
    \caption{Performance evaluation of the HDC encodings, with chronologically ordered data.\label{tab:encoding_selection_o}}

    \begin{tabular}{|l|l|l|l|l|}
    \hline
        \textbf{Metric} & \textbf{KV (RL)} & \textbf{S. Proj} & \textbf{KV (SN)} & \textbf{Generic} \\ \hline
        Accuracy & 89.274 & 89.476 & 84.085 & \textbf{89.47}  \\ \hline
        F1-Score & 0.239 & 0.242 & \textbf{0.287} & 0.241  \\ \hline
    \end{tabular}
\end{table}

We tested the ensemble \textit{Generic} encoding and found it to be effective. However, we also aimed to evaluate its performance using different n-gram sizes as this can greatly impact the algorithm's effectiveness. To achieve this, we experimented with sizes ranging from 2 to 7, as studies have shown that this range tends to yield optimal results~\cite{HDC_classification}. According to Figure~\ref{fig:ngram_size}, we can conclude that the optimal n-gram size achieved is 6, particularly when the input data is shuffled. However, if the data is ordered, it is preferable to use an n-gram of size 2, since it provides the same efficiency while being more efficient and will result in faster training and inference. For the purposes of the next experiments, we use n-gram size 6.

\begin{figure}[htbp!]
\centering
\includegraphics[width=7cm]{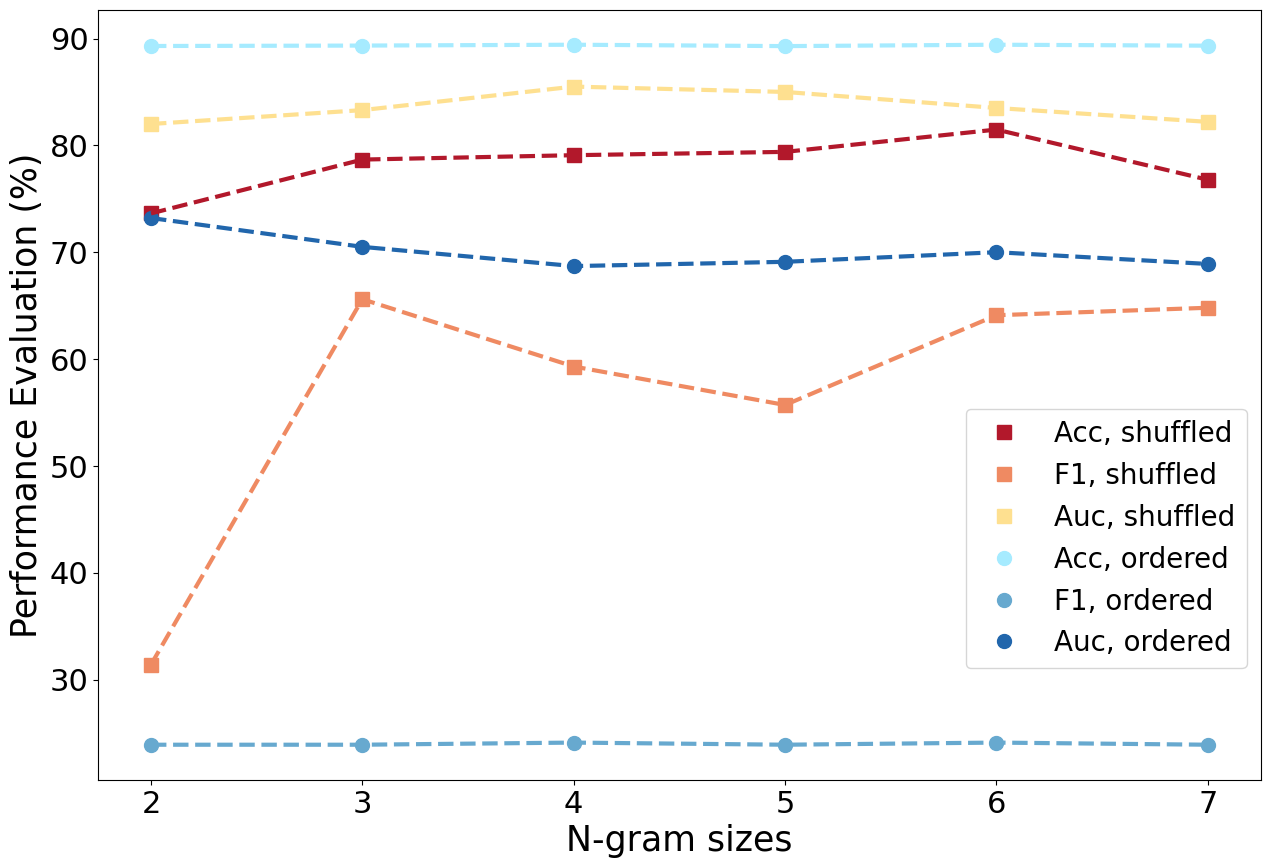}
 \caption{\label{fig:ngram_size} Performance evaluation of the generic encoding with different n-gram sizes, with randomly shuffled and chronologically ordered data.}
\end{figure}

\subsection{Second Phase Results - Learning Method Model Selection}

Combining the ensemble \textit{Generic} encoding with the learning method models described in Section~\ref{HDC Approach}, we again experimented with both randomly shuffled and chronologically ordered data, and collected accuracy and F1-metrics as shown in Table~\ref{tab:method_selection_s} and Table~\ref{tab:method_selection_o}. Using the \textit{RefineHD} model had one of the highest accuracy scores among the different models, for both shuffled and ordered data. Similar to the previous tests, the F1-Score was lower for the ordered data due to the negative/positive label ratio. We decided to use this combination for our experimentation in the upcoming section.

\begin{table}[!ht]
    \centering
    \caption{Performance evaluation of the HDC classification models, with randomly shuffled data.\label{tab:method_selection_s}}

    \begin{tabular}{|l|l|l|l|l|l|l|}
    \hline
        \textbf{Metric} & \textbf{V} & \textbf{A} & \textbf{O} & \textbf{R} & \textbf{N} & \textbf{D} \\ \hline
        Accuracy & 66.63 & 78.62 & 82.62 & \textbf{84.00} & 82.34 & 76.132  \\ \hline
        F1-Score &  0.51 & 0.69 & 0.56  & \textbf{0.70} & 0.70 & 0.66 \\ \hline
    \end{tabular}
\end{table}

\begin{table}[!ht]
    \centering
    \caption{Performance evaluation of the HDC classification models, with chronologically ordered data.\label{tab:method_selection_o}}

    \begin{tabular}{|l|l|l|l|l|l|l|}
    \hline
        \textbf{Metric} & \textbf{V} & \textbf{A} & \textbf{O} & \textbf{R} & \textbf{N} & \textbf{D} \\ \hline
        Accuracy & 47.18 & \textbf{89.73} & 78.75 & 89.47 & 89.69 & 62.94  \\ \hline
        F1-Score & 0.139 & \textbf{0.24} & 0.23 & 0.24 & \textbf{0.24} & 0.14  \\ \hline
    \end{tabular}
\end{table}

\subsection{Third Phase Results - Hyperparameter Tuning}

After deciding on an ensemble \textit{Generic} encoding with an n-gram size of 6 and choosing the HDC learning method model (\textit{RefineHD}), the next important step was to determine the optimal learning rate for the iterative version. We experimented with learning rates ranging from 1 to 5 for both shuffled and ordered data to find the best option. The learning rates for HDC learning method models are typically larger than those used for conventional ML models.


\begin{table}[H]
    \centering
    \caption{Performance evaluation of the ensemble Generic encoding and RefineHD model using different learning rates, with randomly shuffled data.\label{tab:learning_rate_s}
    }
    \begin{tabular}{|l|l|l|l|l|l|l|}
    \hline
        \textbf{Metric} & \boldmath$\alpha=1$ & \boldmath$\alpha=2$ & \boldmath$\alpha=3$ & \boldmath$\alpha=4$ & \boldmath$\alpha=5$ \\ \hline
        Accuracy  & 76.64 & 81.85 & \textbf{82.41} & 81.30 & 78.63  \\ \hline
        F1-Score & 0.659 & 0.637 & \textbf{0.692} & 0.684 & 0.628  \\ \hline
    \end{tabular}
\end{table}

\begin{table}[H]
    \centering
    \caption{Performance evaluation of the ensemble Generic encoding and RefineHD model using different learning rates, with chronologically ordered data.\label{tab:learning_rate_o}
    }
    \begin{tabular}{|l|l|l|l|l|l|l|}
    \hline
        \textbf{Metric} & \boldmath$\alpha=1$ & \boldmath$\alpha=2$ & \boldmath$\alpha=3$ & \boldmath$\alpha=4$ & \boldmath$\alpha=5$ \\ \hline
        Accuracy  & 89.43 & 89.33 & \textbf{89.47} & 89.38 & 89.4  \\ \hline
        F1-Score  & 0.241 & 0.239 & \textbf{0.241} & 0.24 & 0.24  \\ \hline
    \end{tabular}
\end{table}

\begin{figure}[htbp!]
\centering
\includegraphics[width=6cm]{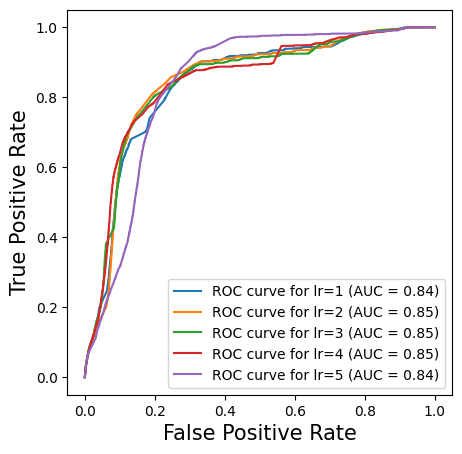}
 \caption{\label{fig:roc_learning_rate} ROC curve evaluation of the ensemble Generic encoding and RefineHD model with different learning rates, with randomly shuffled data.}
\end{figure}

\begin{table}[H]
    \centering
    \caption{Comparison of Precision, recall, and sober/drunk accuracy breakdown with learning rate $\alpha = 3$, with randomly shuffled and chronologically ordered data.\label{tab:sober_drunk_recall}
    }
    \begin{tabular}{|l|l|l|l|l|l|l|}
    \hline
        \textbf{Metric} & \textbf{Randomly Shuffled} & \textbf{Chronologically Ordered} \\ \hline
        Accuracy & 82.41 & 89.47\\ \hline
        Sober Accuracy & 86.12 & 93.33\\ \hline
        Drunk Accuracy & 72.63 & 28.51\\ \hline
        Precision & 0.661 & 0.209\\ \hline
        Recall & 0.726 & 0.285\\ \hline
    \end{tabular}
\end{table}

Upon examining the results presented in Table~\ref{tab:learning_rate_s}, Table~\ref{tab:learning_rate_o}, and the receiver operating characteristic (ROC) curves in Figure~\ref{fig:roc_learning_rate}, it becomes evident that setting the learning rate to 3 yields the highest levels of accuracy with both randomly shuffled data and chronologically ordered data. Looking further into the performance with learning rate $\alpha = 3$ in Table~\ref{tab:sober_drunk_recall}, we consider accuracy, including sober and drunk accuracy, precision, and recall metrics. Although the precision and recall were lower in the chronologically ordered case, due to the negative/positive label ratio, We can see that the classifier excels at the classification of the negative case (sober).

\subsection{Baseline Comparison with Original Study}

As a preliminary step, we sought to replicate the previous study and make a comparative assessment of our approach. In this specific experiment, we compared the results of our HDC approach using randomly shuffled data and chronologically ordered data with results of the methods of the previous study, including their results for their SVM, CNN, and random forest approaches~\cite{Killian2019LearningTD}. Table~\ref{tab:baseline} provides a comprehensive view of the results achieved by both methods. Our approach has higher accuracy scores than previous used models, and also higher recall when the data is randomly shuffled. When the data is chronologically ordered, precision and recall drop as before because of the negative/positive label ratio, but the sober accuracy (negative label) is much higher. Notably, our approach demonstrates a significant improvement when the data is chronologically ordered, boasting a 12\% increase in general accuracy and a 12\% enhancement in sober accuracy compared to the results of the best performing random forest model from the previous study.

\begin{table}[H]
    \centering
    \caption{Performance evaluation of the our HDC approach (randomly shuffled data (S) and chronologically ordered data (O)) vs results of previous study.\label{tab:baseline}
    }
    \begin{tabular}{|l|l|l|l|l|l|}
    \hline
        \textbf{Metric} & \textbf{SVM~\cite{Killian2019LearningTD}} & \textbf{CNN~\cite{Killian2019LearningTD}} & \textbf{RF~\cite{Killian2019LearningTD}} & \textbf{HDC (S)} & \textbf{HDC (O)}  \\ \hline
        Accuracy & 74.67 & 74.27 & 77.48 & 82.39 & \textbf{89.47} \\ \hline
        Sober Acc & 0.812 & 0.789 & 0.815 & 0.860 & \textbf{0.933} \\ \hline
        Precision & 0.636 & 0.636 & \textbf{0.665} & 0.661 & 0.209 \\ \hline
        Recall & 0.622 & 0.656 & 0.698 & \textbf{0.725} & 0.285 \\ \hline
    \end{tabular}
\end{table}

Finally with the final combination of our ensemble \textit{Generic} encoder with n-gram size 6 and a learning rate 3 with the \textit{RefineHD} method we obtain an 89\% accuracy when the data is ordered chronologically which surpasses the previous work by \textbf{12\%}.

\subsection{Hardware Execution Times per Window}

To verify the feasibility of our approach for classifying live data in real-time, we measured the average execution time of classifying 1 10 second window with combinations of the four encodings and the \textit{RefineHD} model on the Raspberri Pi. In order to be feasible, classification for a 10 window needs to be significantly less than 10 seconds. We found that our chosen ensemble \textit{Generic} encoding approach with \textit{RefineHD} took on average 0.34 seconds on the Raspberry Pi, as shown in Table~\ref{tab:execution_times}.

\begin{table}[!ht]
    \centering
    \caption{Encodings execution time per sample.\label{tab:execution_times}}

    \begin{tabular}{|l|l|l|l|l|}
    \hline
        \textbf{Metric} & \textbf{KV (RL)} & \textbf{S. Proj} & \textbf{KV (SN)} & \textbf{Generic} \\ \hline
        Time (s)  & 0.3 & \textbf{0.1} &  0.71 & 0.34  \\ \hline
    \end{tabular}
\end{table}

Our experiments on the Raspberry Pi 4 indicate the potential for implementing our HDC approach on a smartphone, or even a smartphone-smartwatch duo. This paves the way for immediate assessments and JITAIs, allowing smartphones to process sensor data from smartwatches and conduct classification without relying on more computing-intensive resources such as cloud servers. 
\section{Future Work}
To address the shortcomings of using only accelerometer data collected from smartphones, we are expanding our consideration to additional sensor measurements besides accelerometer data, collected by smart watches or other smart wearable devices. These will include heart-rate, gyroscope, temperature, and blood oxygenation level measurements, paired with TAC measurements from a TAC measuring bracelet for correlation. These new measurements have been found to have some correlation to alcohol levels and intoxication~\cite{sensing2022}. We are developing a mobile application for data collection of these measurements from these smart wearables as part of a larger future study. 

Once we have a new dataset collected from smart wearables, our next steps would be running our HDC approach on smart wearable devices with the new dataset to classify TAC measurements, and further refining our approach with the results. Then, we can modify our smart wearable application for reading live data. Our end goal is a complete JITAI solution that can be deployed to consumers, and is both affordable and accessible to the general public, ultimately leading to a more significant impact on promoting healthier drinking habits.
\section{Conclusions}
Alcohol consumption significantly impacts an individual's well-being, with more pronounced consequences when consumed excessively. Various strategies aim to promote responsible drinking habits; among these, JITAIs stand out as a promising approach. To address this issue, we proposed the use of HDC to test the feasibility of IoT-based JITAIs that could address a wide range of alcohol consumption issues. Our research demonstrated that hyperdimensional learning outperforms the traditional random forest model in this context. After exploring various encoding methods and hyperdimensional learning techniques, we found that the optimal configuration involves using a generic encoding with an n-gram size of 6, especially when combined with the RefineHD hyperdimensional learning method.

Our study shows significant improvement with an accuracy rate of 89\%, which is 12\% higher than previous results. Additionally, our work corroborates that organizing samples temporally leads to an additional +7\% accuracy, highlighting the advantageous leverage that HDC gains from time series data. In summary, this enables an accurate JITAI solution that would be widely available to the general public, and would be less invasive and cumbersome and provide less stigma when used in practice.

\balance
\bibliography{bibliography}

\end{document}